\documentclass{bmcart}

\usepackage[utf8]{inputenc} 
\usepackage{amstext}
\usepackage{color}
\usepackage{hyperref}
\usepackage{cleveref}
\usepackage{booktabs}
\usepackage{graphicx}

\startlocaldefs
\endlocaldefs

\begin{document}

\begin{frontmatter}

\begin{fmbox}
\dochead{Research}


\title{Symbolic regression outperforms other models for small data sets}


\author[
   addressref={aff1},
   email={casper.wilstrup@abzu.ai}
]{\inits{CW}\fnm{Casper} \snm{Wilstrup}}
\author[
   addressref={aff1},
   email={jaan.kasak@abzu.ai}
]{\inits{JK}\fnm{Jaan} \snm{Kasak}}


\address[id=aff1]{
  \orgname{Abzu AI}, 
  \street{Orient Plads 1, Mezzanine},                     %
  \postcode{2150}                                
  \city{Nordhavn},                              
  \cny{Denmark}                                    
}


\begin{artnotes}
\end{artnotes}

\end{fmbox}


\begin{abstractbox}

\begin{abstract} 

\parttitle{Background} 
  Machine learning is often applied in health science to obtain predictions and new understandings of complex phenomena and relationships, but an availability of sufficient data for model training is a widespread problem.
  Traditional machine learning techniques, such as random forests and gradient boosting, tend to overfit when working with data sets of only a few hundred observations. 

\parttitle{Results} 
  This study demonstrates that for
  small training sets of 250 observations, symbolic regression generalises better to
  out-of-sample data than traditional machine learning frameworks, as measured by the coefficient of determination $R^2$ on the validation set.
  
\parttitle{Conclusion} 
  In 132 out of 240 cases, symbolic regression achieves a higher $R^2$
  than any of the other models on the out-of-sample data. Furthermore, symbolic regression also
  preserves the interpretability of linear models and decision trees, an added benefit to its superior generalisation. The second best algorithm was found to
  be a random forest, which performs best in 37 of the 240 cases. When restricting the comparison to
  interpretable models, symbolic regression performs best in 184 out of 240 cases.
\end{abstract}




\end{abstractbox}
%

\end{frontmatter}



\section*{Background}
In many research fields, it is common to work with data sets of only a few hundred observations or even fewer.
This is typical in health science fields, where the size of available observational data
can not only be small, but due to the nature of the investigation, it can also be difficult to validate your
hypotheses on new data.
For data like these, statistical methods such as ordinary least squares regression have long been the norm. If the researcher has additional knowledge about the process which generated the data, it is possible to choose a specific model that takes this knowledge into account. An example in the context of health science is survival analysis where the ultimate outcome is censored, in which case the researcher may choose a Cox Proportional Hazards Model \cite{Cox1972}.

In either case, these models suffer from the researcher's need to decide on the actual structure of the model before fitting it to the data. For example, if using a linear model, the researcher has implicitly ruled out any non-linear or interaction effects in the data.
One approach to overcome some of the constraints of a fixed model architecture is to use decision trees. A decision tree can discover and model interaction terms in the data and has the advantage of being as easy to inspect and interpret as a linear model. Decision trees rarely improve the prediction accuracy compared to linear models, however, and they tend to overfit unless restricted in various ways \cite{hastie2009}.

In recent years, it has become increasingly popular to complement statistical analysis with machine learning. For example, random forests \cite{Breiman2001} and gradient boosting \cite{friedman1991} have been widely applied. Such machine learning methods can fit the data more accurately than traditional statistical methods. Compared to statistical methods, they are more prone to overfitting, but this can be mitigated by using a range of regularisation techniques \cite{hastie2009}. These methods work by fitting an ensemble of decision trees. The ensemble can be quite big, encompassing hundreds of distinct decision trees where the final prediction is some aggregation of each individual tree's predictions. While such ensembles can be demonstrated to have strong predictive properties, they are impractical or even impossible to interpret by direct inspection. There exists an ecosystem of tools to analyse these opaque models for feature importance or feature interactions \cite{hastie2009,lundberg2017,ribeiro2016}. Using these ensembles as research tools to increase understanding of the problems under investigation requires that the user be an expert. Thus, we question whether fitting an ensemble model to a data set and obtaining a strong predictive model in itself constitutes a scientific result \cite{shmueli2010}. The model can have useful practical or clinical applications, but it does not explain why the predicted outcome occurs.
In contrast, the fixed mathematical form of a specific parametric model, or the inspectable decision boundaries of a decision tree, provide clear insights to the researcher about what the model has learned.

In the hope of overcoming the limitations of fixed parametric models while preserving their direct interpretability, a family of machine learning techniques known as symbolic regression has been developed. Symbolic regression is a machine learning method that attempts to explain some $Y$ in terms of some $X$ using a mathematical expression composed of a set of basic functions. Traditionally, genetic programming has been used to search the space of all such expressions selectively 
\cite{orzechowski2018,schmidt2009,Vladislavleva2009,Dabhi2011}. Some recent approaches have been more physics-inpsired in their approach \cite{udrescu2020,tailin2019,cranmer2019,cranmer2020,navarro2020,kim2020,liu2020}.

The company Abzu, which employs the authors of this article, has recently developed a symbolic regressor inspired by quantum field theory known as the QLattice \cite{feyn}. The QLattice samples the infinite list of possible mathematical expressions that could link $Y$ to $\mathbf{X}$ as a superposition of a set of spatial paths. These paths are interpreted as mathematical equations by their interactions amongst themselves. An interaction between two spatial paths results in adding a binary mathematical operation to the resulting equation, and a spatial path self-interacting results in a unary operation being added. The QLattice searches this space of all mathematical expressions, including parameters, for the expressions that best explain the output in terms of the input. As the QLattice learns the structure of the problem, the probability fields for these spatial paths are updated such that the best mathematical equation to explain the input data becomes more likely. The result of the search is a list of likely functions $Y=f(\mathbf{X})$ sorted by how well they match a provided set of observations $\mathbf{X}$.
Symbolic regression also differs from random forest and gradient boosting in that it does not fit an ensemble. Instead, it yields the simplest possible mathematical expression that is a candidate for explaining the observed data set. In this study, we compare the generalisation ability of the QLattice-based symbolic regression with interpretable linear models and decision trees, as well as to non-interpetable machine learning models. Such comparisons between interpretable models and machine learning techniques have also been investigated before \cite{couronne2018, oriol2019}. The aim is to show how symbolic regression combines the favourable traits of both interpretable parametric models and the more predictive, but less interpretable, machine learning models.

\section*{Methods}
In this section, we outline our approach to comparing the generalisation ability of the QLattice symbolic regression implementation to a selection of mainstream machine learning methods. Each model's ability to generalise is measured by its $R^2$ score on out-of-sample data after being trained on a small training set of 250 observations. First, we introduce the choice of data sets for this comparison. The choice of models and hyperparameters follows, together with scoring schemes we apply to rank the models under investigation.

\subsection*{Data set selection}
\begin{figure}[ht!]
  \centering
  \includegraphics[width=0.48\textwidth]{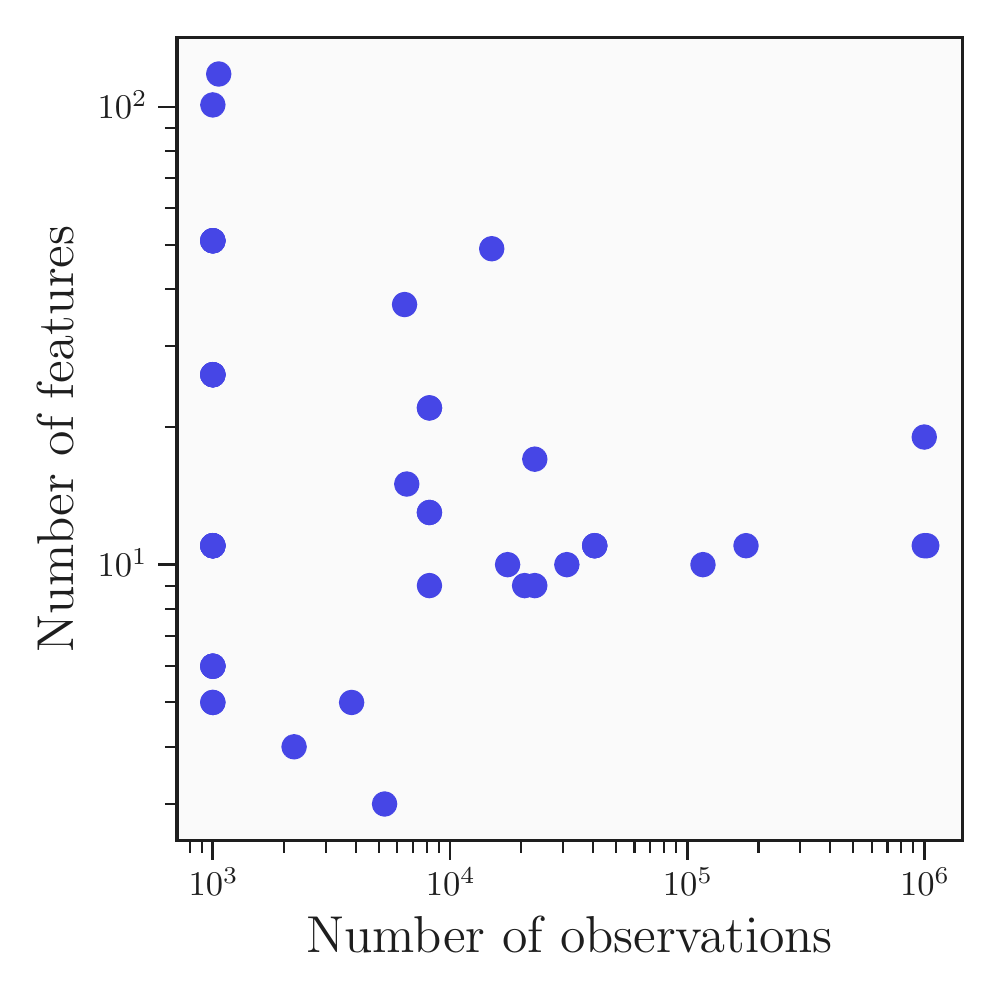}
  \caption{A scatter plot of the 48 data sets used in the comparison. The number of features in each data set is on the $y$ axis, and the number of observations is on the $x$ axis. Both axes use a logarithmic scale.}
  \label{fig:dataset-summary}
\end{figure}

All data sets were taken from the Penn Machine Learning Benchmarks (PMLB)
collection \cite{le2020pmlb}. The data sets were chosen such that a much larger
validation set was available to serve as a measurement of generalisation to
out-of-sample data. In the comparison, we included all regression data sets in
PMLB of at least 1000 observations. This ensures an out-of-sample validation set
of at least 750 observations. The PMLB contains 122 regression data sets, of
which 48 match our criteria. \Cref{fig:dataset-summary} shows the distribution
of features and observations in the selected sets. 
To minimise the effects of possible bias in the choice of training set, we
randomly sampled a training set of 250 observations a total of five times from
each data set, leaving all the remaining observations as out-of-sample
validation data. We therefore had $48 \times 5 = 240$ different data sets on
which to compare the models.

\subsection*{Models in the comparison}
For inspectable models, linear regression and decision trees were chosen. Both are popular because they allow the researcher to interpret and reason about the fitted model directly. We also included two widespread but non-interpretable ensemble learning techniques in the comparison; gradient boosting regression and random forest regression.
For linear regression, decision trees, gradient boosting, and random forests, we used the Scikit-Learn implementations of version 0.24.1 \cite{scikit-learn}. For symbolic regression we used QLattice/Feyn version 1.5.3 \cite{feyn}.

For each of the five model techniques, we chose models with typical hyperparameter settings. For linear regression, we considered ordinary least squares regression a special case of Lasso regression, with a regularisation constant $\alpha = 0$. For decision trees, we used different limits on the depth of the tree. For Gradient Boosting and Random Forest, we used different numbers of estimators in the ensemble. All other parameters were left to the default values in Scikit-learn.

\begin{table}[ht!]
  \centering
  {\footnotesize
  \begin{tabular}{ll}
  \toprule
  Model type & Details / parameters \\

  \hline

  QLattice regression & QLattice(criterion="aic", max\_edges=11) \\
  QLattice regression & QLattice(criterion="bic", max\_edges=11) \\

  \hline

  Linear Regression & LinearRegression() \\
  Lasso & Lasso(alpha=0.01, max\_iter=100000) \\
  Lasso & Lasso(alpha=0.05, max\_iter=100000) \\
  Lasso & Lasso(alpha=0.10, max\_iter=100000) \\

  \hline

  Decision Tree & DecisionTreeRegressor(max\_depth=1) \\
  Decision Tree & DecisionTreeRegressor(max\_depth=2) \\
  Decision Tree & DecisionTreeRegressor(max\_depth=4) \\
  Decision Tree & DecisionTreeRegressor(max\_depth=6) \\

  \hline

  Random Forest & RandomForestRegressor(n\_estimators=400) \\
  Random Forest & RandomForestRegressor(n\_estimators=200) \\
  Random Forest & RandomForestRegressor(n\_estimators=100) \\
  Random Forest & RandomForestRegressor(n\_estimators=50) \\

  \hline

  Gradient Boosting & GradientBoostingRegressor(n\_estimators=400) \\
  Gradient Boosting & GradientBoostingRegressor(n\_estimators=200) \\
  Gradient Boosting & GradientBoostingRegressor(n\_estimators=100) \\
  Gradient Boosting & GradientBoostingRegressor(n\_estimators=50) \\

  \toprule
  \end{tabular}
  }
  \caption{Models and configurations in the comparison. For each framework, four different typical hyperparameter settings were tested, except for the QLattice where both of the built-in sorting criteria were tested.}
  \label{tab:models}
\end{table}

For QLattice symbolic regression, we used the two different information criteria available in the software; Akaike information criterion (AIC, \cite{Akaike1974}) and Bayesian information criterion (BIC, \cite{schwarz1978}). We also used the max\_edges property to limit the complexity of expressions that the QLattice is allowed to explore. As an example,
the equation $f(x,y,z) = x\times(y + \exp(z))$ corresponds to a graph that connects $f(x,y,z)$ to
the inputs $x$, $y$ and $z$ with 6 edges
\footnote{
  Elaboration on constraining a QLattice simulation available at
  \url{https://docs.abzu.ai/docs/guides/essentials/filters.html}.
}.
This restriction was applied to reduce the run-time of the algorithm. All the different models in the comparison are listed in \Cref{tab:models}.

\subsection*{Comparing the models}
In each of the 240 cases, the models were trained on the full 250 training data samples. Then the generalisation performance as measured by $R^2$ on the entire out-of-sample validation data was computed. We applied two different weighted scoring methods to judge the overall performance on the 240 cases.

In the first weighted scoring method, the model that achieved the best result receives one point, and other models receive none. All models are then ranked by the number of first places accumulated; the model with the most first places is ranked first, and so on.
This could potentially penalise strong runner-up models that consistently achieve a close second or third place, so a second weighted scoring method was applied; in each of the 240 cases, 5 points are given to the model with the best $R^2$ score, 4 to the second best, and so on, down to 1 point for the fifth performing model.

\section*{Results}
\begin{table}[ht!]
  \centering
  {\scriptsize
  \begin{tabular}{l p{10mm} p{13mm} p{15mm} p{13mm} }
  \toprule
  model &
  First \newline places &
  Weighted \newline scoring &
  First places \newline for best &
  Weighted \newline scoring for \newline best \\

  \midrule
  QLattice(criterion="bic", max\_edges=11) & 77 & 644 & 132 & 1033 \\
  QLattice(criterion="aic", max\_edges=11) & 65 & 608 & & \\
  Lasso(alpha=0.1, max\_iter=100000) & 18 & 404 & 32 & 511 \\
  GradientBoostingRegressor(n\_estimators=400) & 12 & 375 & 36 & 821 \\
  RandomForestRegressor(n\_estimators=400) & 10 & 268 & 37 & 787 \\
  LinearRegression() & 9 & 170 & & \\
  GradientBoostingRegressor(n\_estimators=50) & 8 & 166 & & \\
  GradientBoostingRegressor(n\_estimators=200) & 7 & 160 & & \\
  GradientBoostingRegressor() & 7 & 158 & & \\
  Lasso(alpha=0.01, max\_iter=100000) & 7 & 133 & & \\
  RandomForestRegressor(n\_estimators=50) & 5 & 128 & & \\
  RandomForestRegressor() & 5 & 124 & & \\
  RandomForestRegressor(n\_estimators=200) & 4 & 124 & & \\
  DecisionTreeRegressor(max\_depth=2) & 3 & 88 & 3 & 448 \\
  Lasso(alpha=0.05, max\_iter=100000) & 2 & 25 & & \\
  DecisionTreeRegressor(max\_depth=1) & 1 & 20 & & \\
  DecisionTreeRegressor(max\_depth=6) & 0 & 4 & & \\
  DecisionTreeRegressor(max\_depth=4) & 0 & 1 & & \\
  \bottomrule
  \end{tabular}
  }
  \caption{
    The summary results of our benchmarking. Each model included in the analysis
    is listed in the leftmost column, together with relevant model parameters. The models are
    listed in descending order by their points in the `First places' column.
    That column lists the number of times each model performed best on the validation set.
    The next column `Weighted scoring' displays the number of points earned
    under the weighted scoring scheme, which was kinder to runner-up models. The third
    column `First places for best' displays the number of first places earned for
    the best configuration in each class of models. The final column `Weighted scoring
    for best' reports the weighted scoring scheme applied to the best performers of
    each model class.
  }
  \label{tab:all_results}
\end{table}
The overall results in \Cref{tab:all_results} showed that the QLattice symbolic regressor outperformed any of the other modelling or machine learning techniques tested. Both the built-in sorting methods used in the QLattice and BIC sorting and QLattice and AIC sorting yielded better results on average across the 240 experiments than either of the alternative model configurations. The results from the `First places' scoring scheme show the number of tests in which each individual model configuration came out as the top performer measured by $R^2$ on the out-of-sample data set. In 77 out of the 240 experiments, the QLattice symbolic regressor with BIC sorting had the best performance on the validation data. The second best model was the QLattice symbolic regressor using the less conservative AIC sorting. The best non-QLattice model was Lasso linear regression with $\alpha=0.1$, with the less regularised lasso models and ordinary least squares performing worse. The poorest performer among the tested models was the decision tree.

These numbers could hide other competitive models whose $R^2$ is only slightly less than the winner's. The second weighted scoring scheme also rewards runners-up as well as the winner, and the results of this comparison are in
the `Weighted scoring' column of \Cref{tab:all_results}. The ranking of models did not change, and once again, the best overall performing model was QLattice symbolic regression using BIC with a total of 644 points. Interestingly, in this comparison, the random forest and gradient boosting models were still not as strong as the lasso models.

\subsection*{Comparing the best performing configuration for each technology}

\begin{figure}[hbt!]
  \centering
  \includegraphics[width=0.95\textwidth]{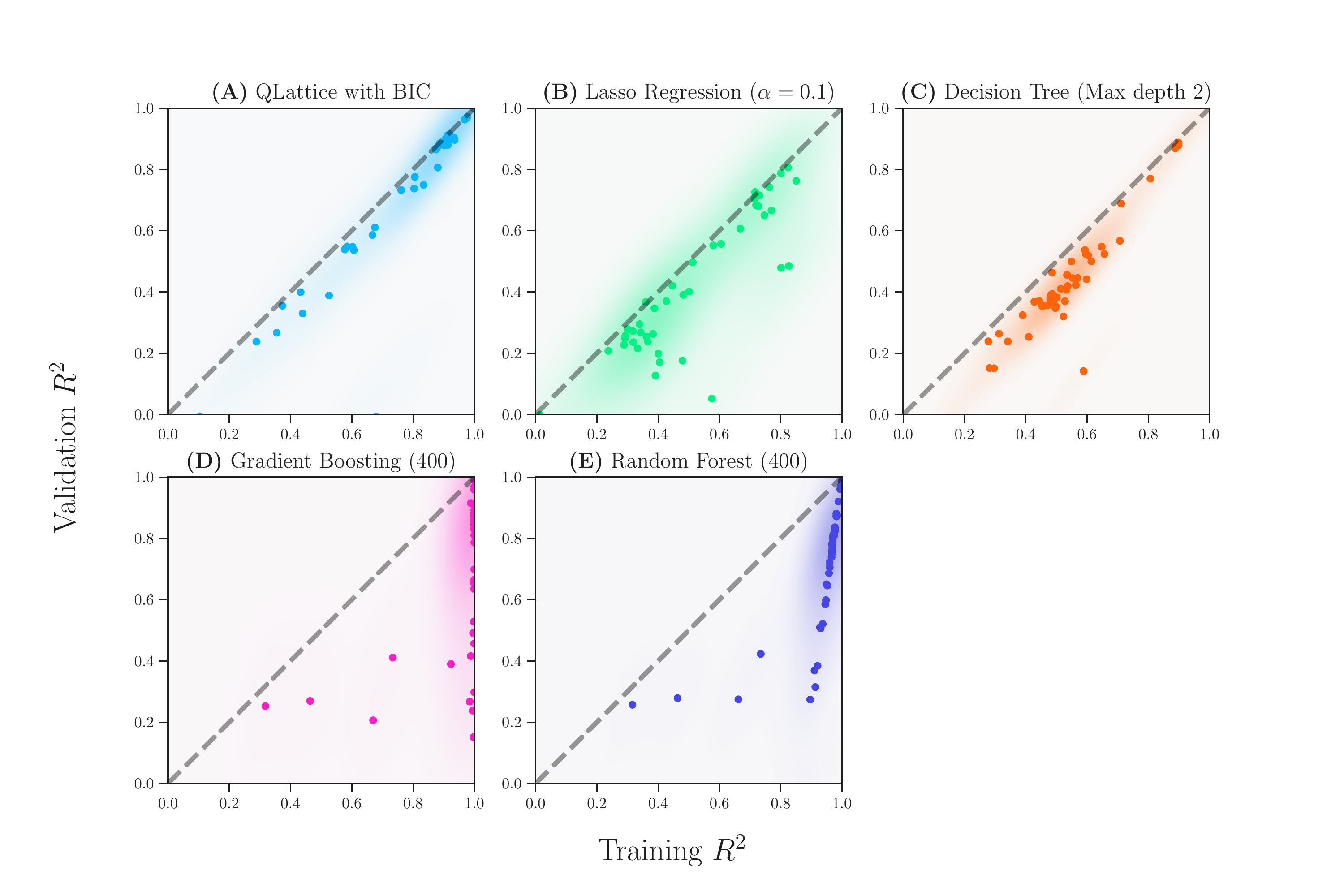}
  \caption{
    Every subplot corresponds to the best generalising models identified from all model
    classes considered, with each point corresponding to the 
    average performance on the five resamplings of each of the 48 underlying data sets,
    as expanded on in the Methods section.
    Each plot also includes a gray dashed line 
    $R^2(\text{training data}) = R^2(\text{validation data})$. A model that is good at
    generalising clusters around this line.
    The first row of plots contains metrics for the models we deem insepctable.
    The second row contains metrics for the ensemble models.
    We call a model good at learning if it has achieved a high $R^2$ on the training set,
    and good at generalising if it achieved a similar $R^2$ on the validation set as on
    the training set.
    We can characterise the five model classes by their performance on our selected data as
    follows. Decision trees with a maximum depth of 2 (C) do not learn well,
    and there is still a small consistent amount overfitting.
    Lasso regression (B) is similar and it generalises slightly better.
    Both gradient boosting (D) and random forest (E) models learn very well,
    but are bad at generalising. 
    Symbolic regression (A) combines these characteristics.
    It is often a good learner, and if not then the model still generalises reliably.
    All plots were bound in the $[0,1]$ range in both axes to preserve the aspect ratio,
    however each model also encountered cases for which $R^2(\text{validation data}) < 0$.
    This happened 4 times for symbolic regression, 3 times for the decision tree,
    2 times for the random forest and gradient boosting, and 5 times for lasso regression.
  }
  \label{fig:scatters}
\end{figure}

As the ranking of models in the `First places' column of \Cref{tab:all_results} shows,
the best performing configuration for each technology was;
\begin{enumerate}
  \item QLattice symbolic regressor with BIC
  \item Lasso linear regressor with $\alpha=0.1$
  \item Gradient boosting with 400 estimators
  \item Random forest regressor with 400 estimators
  \item Decision tree regressor with depth$=2$.
\end{enumerate}
Next we limit our pool of models to these five, and have them compete again for both first places and
under the weighted scoring scheme across the 240 data sets.
The results for these specific configurations are in the last two columns of \Cref{tab:all_results}.
When counting first places, the QLattice BIC model performs best in 132 cases, or slightly more than half of the cases. Here a non-inspectable model first beat the well-performing Lasso regression, with a random forest of 400 estimators scoring 37 points over the regressors 32. In the weighted scoring of these best configurations, gradient boosting and the random
forest finally outperform Lasso regression, with 821 and 787 points compared to the 511 points of Lasso.
This can be explained by the inclusion of only one QLattice regressor as a competitor.
To further analyse the performance and generalisation
of these best performers, we also provide the scatterplot of evaluated $R^2$ scores in \Cref{fig:scatters}.

The benefit of the QLattice was even more apparent when only comparing it to the two modelling technologies which we consider inspectable. When only comparing the QLattice (BIC), Lasso regression with $\alpha=0.1$, and a decision tree regressor of depth 2, the QLattice came out as the best performer in 184 of the 240 cases. Lasso generalised best in 49 cases,
and the simple decision tree was best in 7 cases.

\newpage

\section*{Discussion}
This paper aimed to test the generalisation ability of several machine learning approaches.
There exists a generating distribution $\mathcal{D}$ that a set of observations has been drawn from.
This distribution includes both the predictors and the output, be it a class label or
a continuous variable. In this context, a hypothesis would be a model or explanation
that is a candidate for mapping the predictors to the output. The exercise is to find
a hypothesis that can form good predictions for any new observations drawn from $\mathcal{D}$.
The goodness of predictions is usually measured by prediction accuracy or the
area under curve (AUC) in classification settings and by the $R^2$ score in
regression settings, as is our case.

\subsection*{Occam's Razor}
The crux of the problem is often that the amount of data available from $\mathcal{D}$ is limited, and models that accurately predict on any new out-of-sample observations are hard to create. In our experimental setup, we simulate this challenging scenario by limiting the number of observations available in the training set. Explicitly, we always use training sets of 250 observations, keeping at least 750 others in the holdout set. By keeping the majority of observations in the latter set, we are directly challenging the ability of each candidate to generalise.

We argue that symbolic regression techniques outperform other competitors on the task of generalisation, even when including typically high-performing ensemble models like random forests and gradient boosting as candidates. Regarding the lack of available training data, a key worry of any data scientist using these models should be overfitting. The number of free parameters or model complexity starts playing an important role, and the concise mathematical expressions that symbolic regression realises fight this complexity. This is really just an expression of Occam's Razor in the context of machine learning, where the simpler explanation is the more trustworthy. The trustworthiness comes from the model being easier to reason about and the ability to confirm or invalidate it by applying domain knowledge from your field to the precise dynamics you are modelling. Defending a random forest or another ensemble model from critique about overfitting will always be harder, since your grasp of patterns the model has identified will be poorer.
From the results of our experiment, we see in \Cref{fig:scatters} that the models with simple
structures generalise better; decision trees, linear regression, and symbolic regression
cluster around the
$R^2(\text{training set}) = R^2(\text{validation set})$ line, while the ensemble methods are typically
strong performers on the training set, but not always on the validation set.

\subsection*{Cross-validation}
Cross-validation is a popular tool in the drawers of many data scientists. Traditionally, cross-validation is used to estimate error from model predictions on out-of-sample data \cite{hastie2009}. The most used method is $k$-fold cross-validation, in which the available data is split into $k$ sets. Each of them in turn is used as the holdout set, with the remaining $k-1$ subsets used for training. Predicting on each of the $k$ sets and averaging the errors gives an estimate of the error on new out-of-sample data. $k$-fold cross-validation is also used to select models and tune hyperparameters. One would take all their available data, set aside a validation set, and perform the cross-validation on the training set, selecting the model or hyperparameters that perform the best across the folds.

In our experiment, we did not use cross-validation to select the best possible hyperparameters for each of the non-symbolic techniques. Given our aim of measuring generalisation ability and how we sample 5 different training sets from the overall data, it would not have affected the results significantly. Any cross-validation done to select superior models would only have helped the model to predict samples in the training set, and the ability for such a model to generalise to the out-of-sample holdout set would depend solely on distance between the in-sample and out-of-sample data. On some of the five samplings, cross-validation on the training set might have helped the generalisation, and on some others, it might have been disruptive. It is also generally recommended to perform error estimation using cross-validation only when there is a lack of holdout data available for analysis, something from which our experimental setup does not suffer.

\subsection*{Synthetic Data}
Of note is that in the selected 48 data that match our criteria, 19 were
said to be synthetic data after reaching out to the curators of PMLB for more information.
Specifically, these datasets are from the \href{https://www.openml.org/search?type=data}{OpenML}
repository of data \cite{OpenML2013}, where the metadata describes
them as originating from the Friedman equation \cite{friedman1991}. As specified in the experimental procedures section, our selection criteria enforced nothing about the origin of each available data set, just that each one should have at least 1000 samples.

\section*{Conclusion}
In this paper we compared the predictive performance of the QLattice symbolic regressor to linear models, decision tree regression, random forest regression and gradient boosting regression on 48 data sets from the Penn Machine Learning Benchmark collection. We conclude that the symbolic regression models generalise better to out-of-sample data than any of the other models in the comparison. At the same time, the simple symbolic regression models are intuitive and easy to interpret, which may help researchers gain new insights into the actual causal pathways in their data sets.

\section*{List of abbreviations}

\section*{Declarations}

\subsection*{Ethics approval and consent to participate}
Not applicable.

\subsection*{Consent for publication}
Not applicable.

\subsection*{Availability of data and materials}
All data used in this study is available from the \href{https://epistasislab.github.io/pmlb/}{Penn Machine Leaning Benchmarks package} \cite{le2020pmlb}.
Code to reproduce this study is publicly available under a BSD 3-Clause License at \url{https://github.com/wilstrup/sr-small-data}.
The code needs access to a QLattice to run. Abzu grants access to QLattices for free for non-commercial purposes. If you need help getting access, please contact the authors or the company.

\subsection*{Competing interests}
Both CW and JK are employees of Abzu ApS, the company that developed and owns the QLattice.

\subsection*{Funding}
Not applicable.

\subsection*{Authors' contributions}
CW invested CPU time to conduct the experiment, both authors participated in the subsequent analysis
of results and writing of the article.

\subsection*{Acknowledgements}
Not applicable.


\begin{backmatter}

\bibliographystyle{bmc-mathphys} 
\bibliography{bibliography}      


\end{backmatter}
\end{document}